\def\year{2021}\relax
\documentclass[letterpaper]{article} 
\usepackage{aaai21}  
\usepackage{times}  
\usepackage{helvet} 
\usepackage{courier}  
\usepackage[hyphens]{url}  
\usepackage{graphicx} 
\usepackage{graphicx}  
\usepackage{natbib}  
\usepackage{caption} 
\usepackage[switch]{lineno}
\usepackage{amssymb}
\usepackage{algorithm}
\usepackage[noend]{algpseudocode}
\usepackage{amsmath,amsthm}
\usepackage{multirow}
\usepackage{multicol}
\usepackage{enumitem}
\usepackage{graphicx} 
\usepackage{caption,subcaption}
\usepackage{amsfonts}
\title{Task Aligned Generative Meta-learning for Zero-shot Learning}
\author{
         Zhe Liu,\textsuperscript{\rm 1}\footnote{Equal contribution.}
         Yun Li, \textsuperscript{\rm 1}$^{*}$
         Lina Yao, \textsuperscript{\rm 1}
         Xianzhi Wang, \textsuperscript{\rm 2},
         Guodong Long \textsuperscript{\rm 2}
 }

\affiliations {
     \\
     \textsuperscript{\rm 1} University of New South Wales, Australia \\
     \textsuperscript{\rm 2} University of Technology Sydney, Australia \\
}

\begin{document}

\maketitle

\begin{abstract}
Zero-shot learning (ZSL) refers to the problem of learning to classify instances from the novel classes (unseen) that are absent in the training set (seen). Most ZSL methods infer the correlation between visual features and attributes to train the classifier for unseen classes. However, such models may have a strong bias towards seen classes during training. Meta-learning has been introduced to mitigate the basis, but meta-ZSL methods are inapplicable when tasks used for training are sampled from diverse distributions. In this regard, we propose a novel Task-aligned Generative Meta-learning model for Zero-shot learning (TGMZ). TGMZ mitigates the potentially biased training and enables meta-ZSL to accommodate real-world datasets containing diverse distributions. TGMZ incorporates an attribute-conditioned task-wise distribution alignment network that projects tasks into a unified distribution to deliver an unbiased model. Our comparisons with state-of-the-art algorithms show the improvements of 2.1\%, 3.0\%, 2.5\%, and 7.6\% achieved by TGMZ on AWA1, AWA2, CUB, and aPY datasets, respectively. TGMZ also outperforms competitors by 3.6\% in generalized zero-shot learning (GZSL) setting and 7.9\% in our proposed fusion-ZSL setting.
\end{abstract}

\section{Introduction}
\begin{figure}
    \centering
    \begin{subfigure}{0.95\linewidth}
    \includegraphics[width=\textwidth]{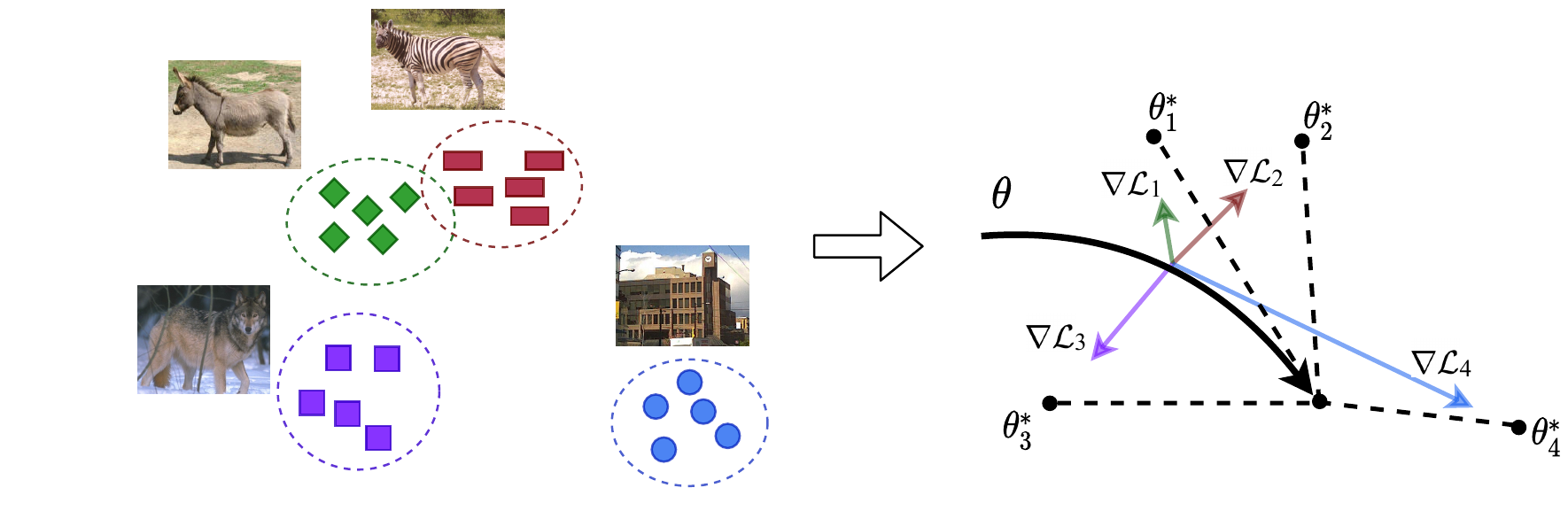}
    \centering
    \caption{Conventional Meta-ZSL.}
    \end{subfigure}
    \begin{subfigure}{0.95\linewidth}
    \includegraphics[width=\textwidth]{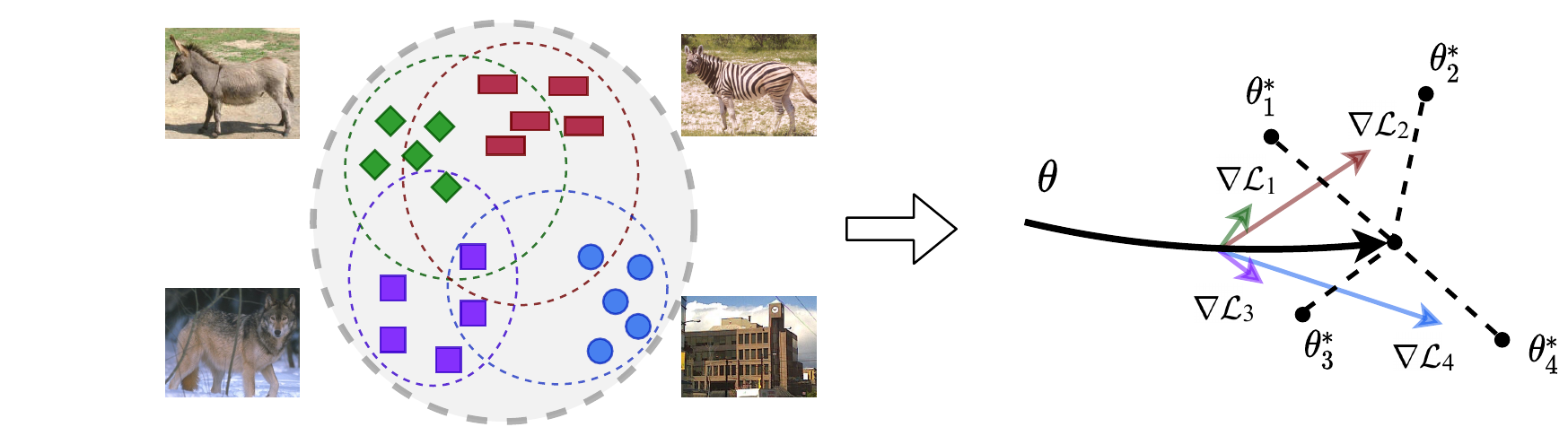}
    \centering
    \caption{Proposed Meta-ZSL.}
    \end{subfigure}
    \caption{Visualization of model representation $\theta$ of the conventional and the proposed meta-ZSL using 1 class for query sets. The conventional meta-ZSL~\cite{finn2017model} is biased towards class 'building', while the proposed aligned task distributions can avoid such local optimality.}
    \label{intro_fig}
\end{figure}


Most current machine learning methods focus on classifying instances into existing classes based on large amounts of labeled data~\cite{day2017survey}. Such methods cannot well handle real-world settings where there are many classes and insufficient instances to cover all the classes~\cite{wang2019survey}. Zero-shot learning (ZSL) aims to classify novel (or unseen) classes based on the existing (or seen) training set and has achieved significant success in many fields, e.g., computer vision~\cite{reed2016learning,zhang2016zero}, natural language processing~\cite{firat2016zero,johnson2017google}, and human activity recognition~\cite{wang2017zero}. 

Typical ZSL algorithms learn the correlation between visual features and the corresponding attributes, e.g., hand-engineering attributes~\cite{shen2018zero,liu2018zero} and textual description~\cite{srivastava2018zero,kumar2019zero}, and utilize the semantic information to infer the classification space for unseen classes. Among ZSL algorithms, attribute-based algorithms classify unseen classes based on visual-attribute embedding~\cite{zhang2018zero,liu2020attribute,Wang_2019_CVPR,al2015transfer}, and generative methods emulate the general data distributions of unseen classes conditioned on attributes and synthesize instances for supervised training of classifiers~\cite{mishra2018generative,zhu2019learning,zhang2018zero}. All the existing ZSL algorithms only optimize model based on seen classes but fail to explicitly mimic ZSL settings that transfer knowledge from seen classes to unseen class at the training time. Consequently, these algorithms are biased towards existing visual-attribute correlation and fail to speculate the real classification space for unseen classes.

To address this issue, some work~\cite{metazeroshot1,meta_zeroshot1,meta_zeroshot2,meta_zeroshot3,meta_zeroshot4,verma2020meta} introduces meta-learning, e.g., model-agnostic meta-learning framework~\cite{finn2017model}, into ZSL, namely meta-ZSL. Meta-ZSL splits existing training classes into two disjoint sets, namely support and query sets, to mimic seen and unseen classes. Then, it randomly picks up classes from support and query sets to construct different tasks for training. This way, meta-ZSL can explicitly learn to adapt from seen classes to unseen classes and thus obtain an unbiased model~\cite{snell2017prototypical,vinyals2016matching}. 

However, current meta-ZSL approaches~\cite{verma2020meta,meta_zeroshot3,meta_zeroshot2} directly integrate meta-learning and ZSL without considering the limitations posed by diverse data distributions in ZSL. Thus, the learned models may be misguided towards extremely different distributions.
Take `wolf', `donkey', `zebra', and `building' from the real-world dataset, aPY~\cite{farhadi2009describing} (shown in Figure \ref{intro_fig}), for example. The first three classes ('wolf', 'donkey', 'zebra') are all animals and dissimilar to the class 'building'. Suppose the four classes are the query sets in four tasks. Conventional meta-ZSL methods would optimize the model representation $\theta$ based on the largest component of the overall gradient of four tasks, i.e., $\bigtriangledown\mathcal{L}_{4}$. Thus, the model representation $\theta$ will be biased towards the optimal solution to 'building' ($\theta^{*}_{4}$) and become less discriminative on animal classes. Therefore, it is necessary to align task distributions in meta-ZSL to enable models to learn each class more moderately and robustly, as illustrated in Figure~\ref{intro_fig} (b).

In this paper, we propose 
a novel Task-aligned Generative Meta-learning model for Zero-shot learning (TGMZ). TGMZ uses attribute-conditioned Task adversarial AutoEncoder (TAE) to align distributions on multiple random tasks with attribute side information. The TAE extracts visual and attribute characteristics from original instances by encoding data into aligned embedding in a unified distribution. Then, a Meta conditional Generative Adversarial Network (MGAN) simulates the unbiased distribution for unseen classes. Each module in MGAN is modified with a meta-learning agent, which optimizes model parameters. To prove the superiority of our idea in handling diverse task distributions, we evaluate our model in the ZSL setting and two more challenging settings: generalized zero-shot learning (GZSL) and our proposed fusion-ZSL setting. GZSL evaluates models on both seen and unseen classes, and the fusion-ZSL setting evaluates models on the fused datasets. Our contributions in this work are summarized as follows:
\begin{itemize}
    \item We propose a novel task-wise alignment generative meta-model, i.e., TGMZ, for zero-shot learning. TGMZ uses attribute-conditioned TAE to align task-wise distributions and adopts MGAN to learn an unbiased model for classifying instances for unseen classes to overcome the potential distribution disjointedness in meta-ZSL.
    \item We carry out extensive ZSL and GZSL experiments on four benchmark datasets. The results exhibit that our model significantly outperforms state-of-the-art algorithms, demonstrating the superiority of TGMZ.
    \item We evaluate our model the effectiveness of TGMZ in handling diverse task distributions under a novel fusion-ZSL setting (i.e., combined dataset experiments). The embedding spaces of the synthetic instances of our model are more discriminative than state-of-the-art algorithms on both single and combined datasets, demonstrating the effectiveness of our task distribution alignment.
\end{itemize}
\section{Related Work}
\subsection{Zero-shot Learning}
The current ZSL algorithms can be divided into two main categories~\cite{mishra2018generative}: attribute-based ZSL and generative ZSL.
Attribute-based algorithms aim to learn the mapping from visual space to the semantic space. They project the instances of unseen classes to attribute embedding and then predict their class labels by finding the most similar class attribute~\cite{romera2015embarrassingly,xian2016latent,zhang2017learning,liu2018generalized}. For example, Kodirov et al.~\cite{kodirov2017semantic} propose to apply an encoder-decoder structure to extract more feature information supervised by reconstruction loss. Changpinyo et al.~\cite{changpinyo2016synthesized} propose to align the semantic space and image space and thus extract more semantic information in the embedding. Zhang et al.~\cite{zhang2018zero} propose a well-established kernel-based method with orthogonality constraints to better learn the non-linear mapping relationship.
Generative ZSL methods classify unseen classes based on synthesizing instances according to attribute information~\cite{verma2017simple,chen2018zero,zhu2018generative,mishra2018generative}. For example, Zhu et al. \cite{zhu2018generative} and Xian et al. \cite{xian2018feature} apply a generative adversarial network~\cite{goodfellow2014generative} with an auxiliary classifier to regularize the generator to carry correct class information based on Fully Connected Networks (FCNs) and Convolutional Neural Networks (CNNs), respectively. Kumar Verma et al. \cite{kumar2018generalized} adopt conditional autoencoder \cite{sohn2015learning} enhanced with a multivariate regressor to achieve generalized synthetic instances. Zhu et al. \cite{zhu2019learning} further propose to optimize the conditional autoencoder with an altering propagation using maximum likelihood estimation.
The above conventional ZSLs only consider the visual-attribute correlation of existing classes and tend to be biased towards the already known attribute mapping or data distribution. 

To better mimic the ZSL setting, some previous work \cite{Wang_2019_CVPR,verma2020meta,metazeroshot1,meta_zeroshot4} introduces meta-learning~\cite{finn2017model} into ZSL to make the model more suitable for transferring knowledge from seen classes to unseen classes.
For example, Wang et al. \cite{Wang_2019_CVPR} fuse meta-learning and attribute-based ZSL method to map visual features to task-aware embedding.
Soh et al.~\cite{meta_zeroshot2} combine meta-learning with CNNs and utilize a single gradient update to obtain a generic initialization suitable for internal learning.
Verma et al.~\cite{verma2020meta} first introduce meta-learning-based generative ZSL and apply meta-learner on each module of generative ZSL.
All the above methods directly combine meta-learning with ZSL while neglecting the bias of meta-learning caused by the mismatch and the diversity of task distributions.

\subsection{Domain Alignment}
Most current domain adaptation works focus on single-source source-to-target alignment~\cite{wilson2020survey,task_align3,task_align2,task_align1}
to handle different task distributions.
For example, Gholami et al. \cite{task_align2} use an adversarial autoencoder and a discriminative discrepancy loss function to align two domains.
Guo et al. \cite{task_align_new1} extend single-source domain adaptation to multiple sources based on the distance discrepancy.
Zhao et al. \cite{task_align_new2} propose end-to-end adversarial domain adaptation for multiple sources based on a pixel-level cycle-consistency loss.
Wang et al. \cite{task_align_new3} enhance multi-source adversarial alignment by introducing task-specific decision boundaries.
The aforementioned work has not considered aligning domains using the attribute side information on multiple sources without a fixed source-to-target relationship.

\subsection{Summary}
Compared with the related work, our contributions are three-fold.
First, we propose task-wise TAE, which extracts both class and visual feature information for reconstruction, to carry out task-wise distribution alignment.
Second, we fuse meta-learning and ZSL in a robust way. TAE uses attribute side information to provide aligned embedding for ZSL and to prevent meta-learner from optimizing the model to be biased due to disjoint task distributions.
Third and the last, differing from previous work that directly learns visual features for classification, our model learns unbiased synthesized embedding, which is more suitable for novel classes.
\begin{figure*}
    \centering
    \includegraphics[width=\linewidth]{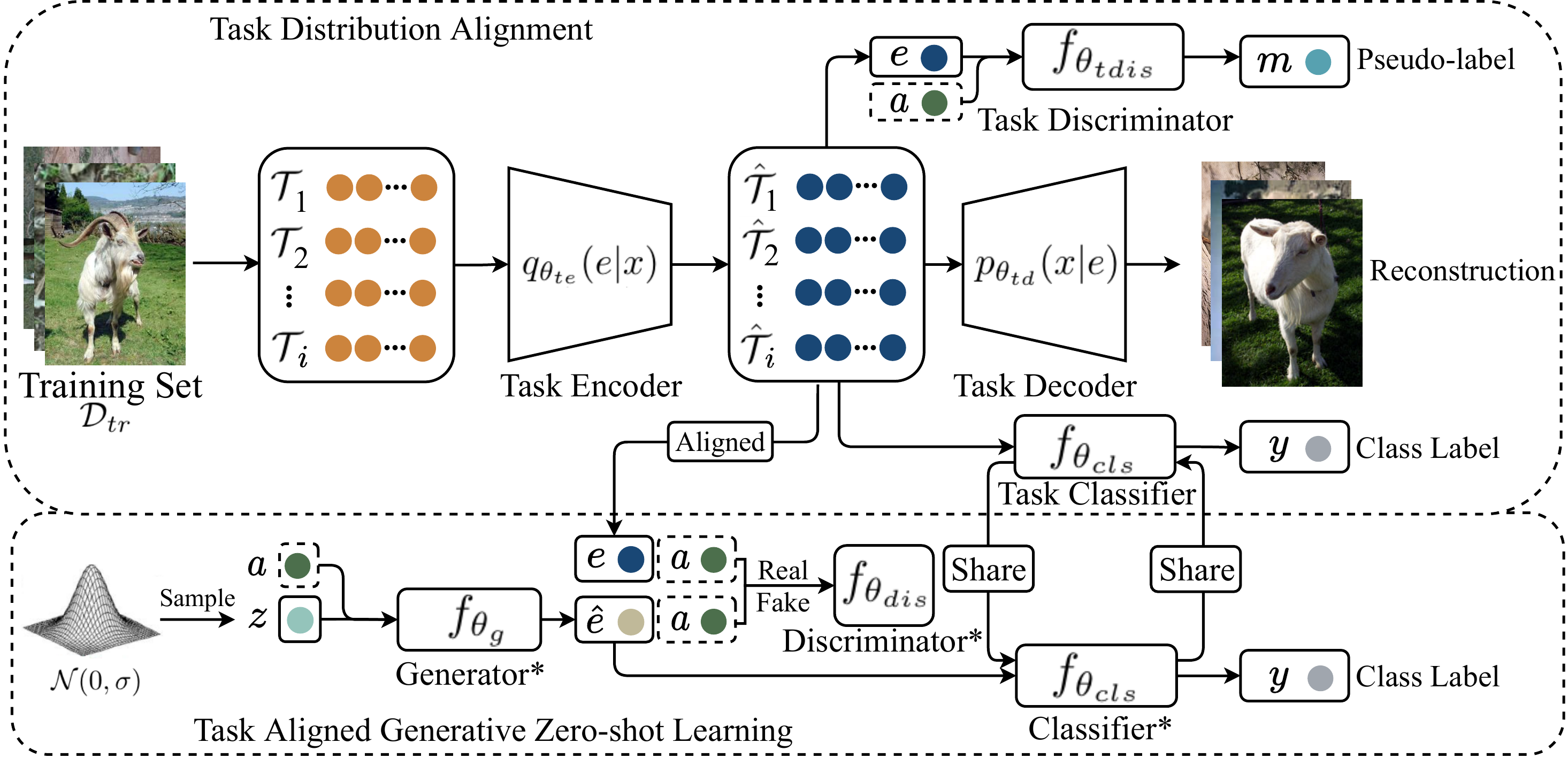}
    \caption{Architecture of the proposed TGMZ. $a$ is the attribute vector. Discriminator$*$, Generator$*$ and Classifier$*$ denote the modules that are updated by meta-learner.}
    \label{fig:my_label}
\end{figure*}
\section{Methodology}
\subsection{Problem Definition and Overview}
Suppose dataset $\mathcal{D}=\{X,A,Y\}$ contains visual features $X$, attribute vectors $A$, and class labels $Y$, and
lowercase letters $x,a,y$ are instances from the respective sets.
$\mathcal{D}$ contains two disjoint subsets: training set $\mathcal{D}_{tr}$ and testing set $\mathcal{D}_{ts}$.
The goal of ZSL is to transfer knowledge from $\mathcal{D}_{tr}$ to $\mathcal{D}_{ts}$.
Our model aims to emulate the attribute-conditioned data distribution to synthesize unseen classes' instances and then train a classifier supervised by the synthetic instances to predict real instances from unseen classes.

Different from conventional ZSL settings, we divide the training set $\mathcal{D}_{tr}$ into a support set $\mathcal{D}_{sup}$ and a disjoint query set $\mathcal{D}_{qry}$ to mimic seen and unseen classes at the training time, respectively. To carry out episode-wise training, we sample tasks $\{\mathcal{T}_{1},\mathcal{T}_{2},...,\mathcal{T}_{i}\}$ following the N-way K-shot setting in ~\cite{verma2020meta}. Given an arbitrary task $\mathcal{T}_{i}=\{\mathcal{D}^{\mathcal{T}_{i}}_{sup},\mathcal{D}^{\mathcal{T}_{i}}_{qry}\}$, we sample $K$ classes with $N$ images for each class from $\mathcal{D}_{sup}$ and $\mathcal{D}_{qry}$, respectively. Our method is inductive~\cite{xian2019zero} so that no extra information from $\mathcal{D}_{ts}$ is used in the training.

Our training procedure consists of two phases: task distribution alignment and task-aligned generative zero-shot learning. The former phase aligns the diverse sampled tasks to a unified task distribution to ease the biased optimization. The latter phase synthesizes instances in the aligned distribution conditioned on attribute vectors and uses meta-learner to train the model. We illustrate the details of the two phases in the following subsections.

\subsection{Task Distribution Alignment}
In this phase, we regularize task distributions into a unified distribution along with the training of TGMZ. We consider three requirements for TAE: (i) the encoder of TAE should be capable of aligning task-wise distributions; (ii) TAE should regularize encoder based on the visual-attribute correlation; (iii) the embedding should extract visual and class characteristics from the original features. 

TGMZ relies on episode-wise training~\cite{Wang_2019_CVPR,verma2020meta} to jointly handle multiple tasks in each iteration while optimizing the model. The first requirement enables the encoder to align multiple sampled task distributions synchronously and to be optimized along with generative networks in the second phase. The second requirement enables TAE to utilize attribute side information during regularization, thus enriching the attribute information in the embedding to better fits the ZSL setting. The third requirement ensures optimizing embedding based on the original visual and class information, which can increase the authenticity of the synthetic instances.

Our proposed TAE consists of three components: Task Encoder $q_{\theta_{te}}(e|x)$, Task Decoder $p_{\theta_{td}}(x|e)$, Task Discriminator $f_{\theta_{tdis}}(e,a)\rightarrow m$, and Task Classifier $f_{\theta_{cls}}(e)\rightarrow y$, where $e$ denotes the encoded visual features, i.e., aligned embedding for task distributions; $x,a,y$ denote the original visual features, attribute vectors and class labels; $\theta$ denotes the parameter for the corresponding module.
The task encoder aims to project tasks into the embedding $e$ following a unified distribution; the task decoder helps reconstruct original instances; task classifier tries to predict the class labels. Specifically, we adopt a multi-label classifier as the discriminator $f_{\theta_{tdis}}(e,a)\rightarrow m$ for TAE and assign pseudo-labels $m$ for tasks to differentiate task distributions, i.e., $\forall \mathcal{T}_{i}$, $m_{\mathcal{T}_{i}}=i$.

For each task $\mathcal{T}_{i}$, we design loss functions $\mathcal{L}_{tdis}$ and $\mathcal{L}_{tc}$ to optimize $\theta_{tdis}$ and other modules $\theta_{tc}$, respectively:
\begin{equation}\label{loss_tc}
\begin{gathered}
\min_{\theta_{tc}}\mathcal{L}_{tc}=\mathbb{E}_{e_{i}\sim q_{\theta_{te}(e_{i}|x_{i})}}[-\log p_{\theta_{td}}(x_{i}|e_{i})]\\
-\mathcal{L}_{ce}(f_{\theta_{tdis}}(e_{i},a_{i}),m_{i})+\mathcal{L}_{ce}(f_{\theta_{cls}}(e_{i}),y_{i})
\end{gathered}
\end{equation}

\begin{equation}\label{loss_tdis}
\begin{gathered}
  \max_{\theta_{tdis}}\mathcal{L}_{tdis}=-\mathcal{L}_{ce}(f_{\theta_{tdis}}(e_{i},a_{i}),m_{i})\\
 s.t. \quad x_{i}\sim \mathcal{T}_{i},\quad \theta_{tc}=\{\theta_{te},\theta_{td},\theta_{cls}\}
\end{gathered}
\end{equation}
where $T_{i}$ denotes a sampled task; $x_{i}$ is an arbitrary sample from $T_i$; $e_{i},a_{i},m_{i},y_{i}$ are the corresponding encoded embedding, attribute vector, pseudo-label, and ground-truth class label for instance $x_{i}$, respectively; $\mathcal{L}_{ce}$ denotes the Cross Entropy loss function.

The second item in $\mathcal{L}_{tc}$ and the loss $\mathcal{L}_{tdis}$ construct a two-player minimax game between the task discriminator and the task encoder. They jointly enable the model to meet the first two requirements.
On the one hand, the task encoder learns to encode diverse tasks into the embedding to confuse the multi-label discriminator with the pseudo-labels $m$;
on the other hand, the task discriminator attempts to distinguish different task distributions.
After such adversarial training, the task encoder will learn to align multiple diverse tasks into a unified distribution to fool task discriminator---the tasks that share the similar distribution will easily confuse task discriminator, and the unique tasks that follow the different distributions will be aligned into a unified distribution, making TAE satisfy the first requirement. Also, the alignment within tasks can also be carried out when tasks contains multiple classes. Since task encoder learns to make different tasks be similar distributions, we can construct different tasks to align the classes within tasks. For example, given four classes $(a,b,c,d)$, if each task contains two classes, we may construct task pairs for episodes by $\{(a,b),(c,d)\}$, and $\{(a,c),(b,d)\}$. During training, task alignment module will learn to align distributions $(a,b)$ to $(c,d)$, $(a,c)$ to $(b,d)$. Finally, after enough episodes, $(a,b,c,d)$ will be aligned in a unified distribution.

Besides, the attribute-conditioned discriminator distinguishes the synthetic instances based on attribute vectors. Since attribute vectors are invisible to the task encoder, the task encoder will be regularized to infer the attribute information of classes from visual features to meet the second requirement.
The other two terms in $\mathcal{L}_{tc}$ are the reconstruction loss and auxiliary classification loss. The former forces the encoder-decoder to reconstruct original visual features, and the latter makes the task encoder extract class information during the encoding, which fulfills the third requirement.

We summarize the episodic loss function $\mathcal{L}_{align}$ for task alignment on multiple tasks $\{\mathcal{T}_{1},\mathcal{T}_{2},...,\mathcal{T}_{i}\}$ as follows:
\begin{equation}\label{loss_alignment}
\begin{gathered}
 \min_{\theta_{tc}}\max_{\theta_{tdis}}\sum_{j=1}^{i}\mathbb{E}_{e_{j}\sim q_{\theta_{te}(e_{j}|x_{j}),x_{j}\sim \mathcal{T}_{j}}}[-\log p_{\theta_{td}}(x_{j}|e_{j})]\\
 -\mathcal{L}_{ce}(f_{\theta_{tdis}}(e_{j},a_{j}),m_{j})+\mathcal{L}_{ce}(f_{\theta_{cls}}(e_{j}),y_{j})
\end{gathered}
\end{equation}

\subsection{Task Aligned Generative Zero-shot Learning}
We learn a general data distribution to generate instances for unseen classes using MGAN. Specifically, we adopt the encoded embedding of original tasks, i.e., aligned tasks, as input for MGAN.
Let $\hat{\mathcal{T}}_{i}=\{\hat{\mathcal{T}}_{i}^{sup},\hat{\mathcal{T}}_{i}^{qry}\}$ be the encoded embedding of $\mathcal{T}_{i}$ and the subsets, $\hat{\mathcal{T}}_{i}^{sup}$ and $\hat{\mathcal{T}}_{i}^{qry}$, be the embedding of $\mathcal{D}^{\mathcal{T}_{i}}_{sup},\mathcal{D}^{\mathcal{T}_{i}}_{qry}$, respectively.
Suppose the aligned tasks follows a unified distribution $\hat{\mathcal{T}}_{i}\sim p(\hat{\mathcal{T}})$, and $p(\hat{\mathcal{T}})$ represents the task distribution over the support and query sets.
For MGAN, we apply a Generator $f_{\theta_{g}}(z,a)\rightarrow\hat{e}$, a Discriminator $f_{\theta_{dis}}(e,a)\rightarrow [0,1]$ and a Classifier $f_{\theta_{cls}}(e)\rightarrow y$, where $z$ is the random noise from a normal distribution and $\hat{e}$ denotes synthetic instances. The discriminator predicts real instances as 1 and fake instances as 0. The classifier in MGAN shares the same weights as the task classifier in TAE, which provides a warm-start initialization for training (we use the same notation for the classifiers).
Each module of MGAN is integrated with a meta-leaner, and the training is a two-step procedure based on gradient descent optimization.

MGAN follows a two-step procedure.
First, meta-learner computes task-specific optimal parameters based on the support sets, i.e., $\hat{\mathcal{T}}_{i}^{sup}$ without updating model parameters $\theta$. Differing from conventional meta-learning, we seek the overall optimal parameters for all the tasks in support sets rather than searching a set of optimal model parameters for each task~\cite{verma2020meta} to achieve better stability in optimizing generative models.
Second, meta-learner computes the gradients of optimizing the overall optimal parameters towards query sets, i.e., $\hat{\mathcal{T}}_{i}^{qry}$, and then summarizes the gradients to update model parameters $\theta$; this enables the model to learn transferable parameters from seen classes to unseen classes.

The training relies on the task-specific loss function $\mathcal{L}_{zsl}$ to compute gradients on support and query sets for obtaining the overall optimal parameters.
Therefore, we start by introducing the task-specific loss function $\mathcal{L}_{zsl}$ as follows:
\begin{equation}\label{loss_zsl}
\begin{gathered}
    \min_{\theta_{gc}}\max_{\theta_{dis}}\mathbb{E}_{\hat{\mathcal{T}}_{i}}[f_{\theta_{dis}}(e_{i},a_{i})]-\mathbb{E}_{a_{i},z\sim \mathcal{N}(0,\sigma)} [f_{\theta_{dis}}(\hat{e}_{i},a_{i})]\\
    +\mathcal{L}_{ce}(f_{\theta_{cls}}(\hat{e}_{i}),y_{i})\\
    s.t. \quad \hat{e}_{i}\sim f_{\theta_{g}}(z,a_{i}),\quad \theta_{gc}=\{\theta_{g},\theta_{cls}\}
\end{gathered}
\end{equation}
where $\hat{e}_{i}$ is a synthetic instance conditioned on $a_{i}$; $z$ is the random noise from Normal Distribution $\mathcal{N}(0,\sigma)$; $\mathcal{L}_{ce}$ is the Cross Entropy loss function.

\begin{algorithm}[t]
  \caption{TGMZ Training Procedure}\label{algorithm_process}
  \begin{algorithmic}[1]
   \Require $\mathcal{D}_{tr}$: training dataset \Require $lr_{tc}, lr_{tdis}$: learning rates
   \Require $\alpha_{1},\alpha_{2},\beta_{1},\beta_{2}$: step sizes 
   \State Initialize $\theta_{tc},\theta_{tdis},\theta_{gc},\theta_{dis}$
   \State Split $\mathcal{D}_{tr}$ into disjoint subsets $\mathcal{D}_{sup}$ and $\mathcal{D}_{qry}$
   \While {not done}
   \State Sample tasks $\{\mathcal{T}_{1},\mathcal{T}_{2},...,\mathcal{T}_{i}\}$ from $\mathcal{D}_{sup}$ and $\mathcal{D}_{qry}$
   \State Update $\theta_{tdis}\leftarrow Adam(\mathcal{L}_{tdis},lr_{tdis})$
   \State Update $\theta_{tc}\leftarrow Adam(\mathcal{L}_{tc},lr_{tc})$
   \For{$j\in [1,i]$}
   \State $\hat{\mathcal{T}}_{j}\leftarrow f_{\theta_{te}}(\mathcal{T}_{j})$
   \State Evaluate $\bigtriangledown_{\theta_{dis}}\mathcal{L}_{\hat{\mathcal{T}}^{sup}_{j}}^{zsl}(\theta_{dis})$ w.r.t $\hat{\mathcal{T}}_{j}^{sup}\in \hat{\mathcal{T}}_{j}$
   \State Update $\theta_{dis}^{'}=\theta_{dis}+\alpha_{1}\bigtriangledown_{\theta_{dis}}\mathcal{L}_{\hat{\mathcal{T}}^{sup}_{j}}^{zsl}(\theta_{dis})$
   \State Evaluate $\bigtriangledown_{\theta_{gc}}\mathcal{L}_{\hat{\mathcal{T}}^{sup}_{j}}^{zsl}(\theta_{gc})$ w.r.t $\hat{\mathcal{T}}_{j}^{sup}\in \hat{\mathcal{T}}_{j}$
   \State Update $\theta_{gc}^{'}=\theta_{gc}-\alpha_{2}\bigtriangledown_{\theta_{gc}}\mathcal{L}_{\hat{\mathcal{T}}^{sup}_{j}}^{zsl}(\theta_{gc})$
  \EndFor
  \State Update $\theta_{dis}\leftarrow\theta_{dis}+\beta_{1}\sum_{\hat{\mathcal{T}}_{j}}\bigtriangledown_{\theta_{dis}}\mathcal{L}_{\hat{\mathcal{T}}^{qry}_{j}}^{zsl}(\theta^{'}_{dis})$
  \State Update $\theta_{gc}\leftarrow\theta_{gc}-\beta_{2}\sum_{\hat{\mathcal{T}}_{j}}\bigtriangledown_{\theta_{gc}}\mathcal{L}_{\hat{\mathcal{T}}^{qry}_{j}}^{zsl}(\theta^{'}_{gc})$
\EndWhile
  \end{algorithmic}
\end{algorithm}

We optimize the task-specific loss function in an adversarial manner.
In each episode, $\mathcal{L}_{zsl}$ fist optimizes the discriminator $\theta_{dis}$ based on the first two items to enables the discriminator to distinguish real and synthetic instances precisely.
Then, $\mathcal{L}_{zsl}$ optimizes the generator and the classifier, i.e., $\theta_{gc}$, to synthesize instances with class information and to confuse attribute-conditioned discriminator according to the last two items.
Finally, the generator emulates the real data distribution based on attribute vectors for unseen classes. 

Let $\bigtriangledown_{\theta_{dis}}\mathcal{L}_{\hat{T}_{i}}^{zsl}(\theta_{dis})$ be the gradient of $\mathcal{L}_{zsl}$ for module $\theta_{dis}$ (subscript of $\bigtriangledown$) conditioned on parameter $\theta_{dis}$ (in brackets) and data $\hat{\mathcal{T}}_{i}$. Similarly, we signify $\bigtriangledown_{\theta_{gc}}\mathcal{L}_{\hat{T}_{i}}^{zsl}(\theta_{gc})$ as the gradient for $\theta_{gc}$ under the same condition. On this basis, we sort out the process of finding the overall task-specific optimal parameters for $\hat{\mathcal{T}}$ as follows:
\begin{equation}\label{loss_first_step_dis}
    \theta_{dis}^{'}=\theta_{dis}+\alpha_{1}\bigtriangledown_{\theta_{dis}}\sum_{\hat{\mathcal{T}}_{j}\sim p(\hat{T})}\mathcal{L}_{\hat{\mathcal{T}}^{sup}_{j}}^{zsl}(\theta_{dis})
\end{equation}
\begin{equation}\label{loss_first_step_gc}
\theta_{gc}^{'}=\theta_{gc}-\alpha_{2}\bigtriangledown_{\theta_{gc}}\sum_{\hat{\mathcal{T}}_{j}\sim p(\hat{T})}\mathcal{L}_{\hat{\mathcal{T}}^{sup}_{j}}^{zsl}(\theta_{gc})
\end{equation}
where $\alpha_{1},\alpha_{2}$ denote the step sizes for the optimization.

With optimal parameters for seen classes of sampled tasks, the meta-learner updates each module as follows:
\begin{equation}\label{loss_second_step_dis}
\theta_{dis}\leftarrow\theta_{dis}+\beta_{1}\bigtriangledown_{\theta_{dis}}\sum_{\hat{\mathcal{T}}_{j}\sim p(\hat{\mathcal{T}})}\mathcal{L}_{\hat{\mathcal{T}}^{qry}_{j}}^{zsl}(\theta^{'}_{dis})
\end{equation}
\begin{equation}\label{loss_second_step_gc}
\theta_{gc}^{'}\leftarrow\theta_{gc}-\beta_{2}\bigtriangledown_{\theta_{gc}}\sum_{\hat{\mathcal{T}}_{j}\sim p(\hat{\mathcal{T}})}\mathcal{L}_{\hat{\mathcal{T}}^{qry}_{j}}^{zsl}(\theta^{'}_{gc})
\end{equation}
where $\beta_{1},\beta_{2}$ denote step sizes for transfer learning from seen classes to unseen classes.

Eq.~\ref{loss_second_step_dis} and Eq.~\ref{loss_second_step_gc} illustrate the episode-wise optimization for TGMZ, with the detailed algorithm procedure described in Algorithm~\ref{algorithm_process}.

\begin{table}[t]
    \centering
    \small
    \begin{tabular}{|c|ccc|}
        \hline
        Dataset & \#Attribute Dim& \#Image & \#Seen/Unseen\\
        \hline
         AWA1 & 85 & 30,475 & 40/10\\
         AWA2 & 85 & 37.322 & 40/10\\
         CUB & 1024 & 11,788 & 150/50\\
         aPY & 64 & 15,339 & 20/12\\
         \hline
    \end{tabular}
    \caption{Dataset statistics. \# denotes number.}
    \label{dataset_statistics}
\end{table}

\begin{table*}[h]
\centering
\small
\caption{ZSL average per-class Top-1 accuracy results. Attribute-based methods are shown at the top and generative methods are at the bottom. * denotes meta-ZSL method. $^{\psi}$ denotes using CNN-RNN feature for CUB dataset.}
\label{ZSL_experiment}
\begin{tabular}{l|c|c|c|c}
\hline 
Method & ~~~~~AWA2~~~~~& ~~~~~AWA1~~~~~ & ~~~~~CUB~~~~~ & ~~~~~aPY~~~~~ \\ 
\hline
\hline 
$^{\psi}$ESZSL~\cite{romera2015embarrassingly}~~~~ & 58.6 & 58.2 & 53.9 & 38.3 \\
$^{\psi}$LATEM~\cite{xian2016latent}~~~~ & 55.8 & 55.1 & 49.3 & 35.2 \\
$^{\psi}$SYNC~\cite{changpinyo2016synthesized}~~~~& 46.6 & 54.0 & 55.6 & 23.9 \\
$^{\psi}$DEM~\cite{zhang2017learning}~~~~ & 67.1  & 68.4 & 51.7 & 35.0 \\
SAE~\cite{kodirov2017semantic}~~~~ & 54.1 & 53.0 & 33.3 & 24.1\\
Gaussian-Kernal~\cite{zhang2018zero}~~~~ & 61.6 & 60.5 & 52.2 &38.9\\
*TAFE-Net~\cite{Wang_2019_CVPR}~~~~ & 69.3 & 70.8 & 56.9 & 42.2 \\
APNet~\cite{liu2020attribute}~~~~  & 68.0 & 68.0 & 57.7 & 41.3 \\
\hline
\hline
$^{\psi}$GFZSL~\cite{verma2017simple}~~~~ & 67.0  & 69.4 & 49.2 & 38.4 \\
$^{\psi}$SP-AEN~\cite{chen2018zero}~~~~ & 58.5 & - & 55.4 & 24.1 \\
GAZSL~\cite{zhu2018generative}~~~~ & 70.2 & 68.2 & 55.8 & 41.3 \\
$^{\psi}$SE-GZSL~\cite{kumar2018generalized}~~~~  & 69.2 & 69.5 & 59.6 & - \\
ABP~\cite{zhu2019learning}~~~~ & 70.4 & 69.3 & 58.5 & -\\
ZVAE~\cite{gao2020zero}~~~~ & 69.3 & 71.4 & 54.8 & 37.4 \\
$^{\psi}$*ZSML~\cite{verma2020meta}~~~~  & 76.1 & 73.5 & 68.3 & 35.0 \\
\hline
\hline
$^{\psi}$*TGMZ-SVM~~~~  & 73.2& 70.9 & \textbf{70.0} & 44.6\\
$^{\psi}$*TGMZ-Softmax~~~~  & \textbf{78.4} & \textbf{75.1} & 66.1 & \textbf{45.4}\\
\hline
\end{tabular}
\end{table*}

\section{Experiment}
\subsection{Experiment Setup}
We conduct extensive experiments on four benchmark datasets: AWA1~\cite{lampert2009learning}, AWA2~\cite{xian2019zero}, CUB~\cite{welinder2010caltech}, and aPY~\cite{farhadi2009describing}.
AWA1, AWA2, and CUB are animal datasets. In particular, CUB consists of fine-grained bird species that are hard to discriminate; aPY comprises highly diverse classes, e.g., buildings and animals. We use hand-engineering attribute vectors in AWA1, AWA2, and aPY, and use 1024-dimensional embedding attributes extracted by Reed et al.~\cite{reed2016learning} in the CUB dataset, which shows superior performance than the original attributes.
We divide the datasets into seen and unseen classes following the proposed split (PS)~\cite{xian2019zero} and adopt visual features from pre-trained ResNet-101, according to Xian et al.~\cite{xian2019zero}. The dataset statistics and train/test split are shown in Table \ref{dataset_statistics}.

We compare our method with fifteen state-of-the-art algorithms in ZSL and GZSL and four representative algorithms in fusion-ZSL. Note that we provide the reproduced ZSML, GZSL, ABP, and DEM results in our experiments. In the ZSL and fusion-ZSL settings, we evaluate our model using Linear-SVM~\cite{verma2020meta} and Softmax (i.e., two fully connected layers followed by batch normalization). Since SVM is time-consuming, we only use Softmax as the classifier to evaluate our model in the GZSL setting. More details about \textit{Dataset Description}, \textit{Model Architecture}, \textit{Parameter Setting} and \textit{Convergence Analysis} can be found in \textbf{Supplementary Material}.

\subsection{Zero-shot Learning}

In the ZSL setting, we evaluate our model using linear-SVM and Softmax (Table \ref{ZSL_experiment}). Compared with extensive state-of-the-art algorithms, our model achieves 2.1\%, 3.0\%, 2.5\% and 7.6\% relative improvements
on AWA1, AWA2, CUB, and aPY, respectively.
Both SVM and Softmax can achieve state-of-the-art performance, with TGMZ-Softmax outperforming the other algorithms on three datasets and TGMZ-SVM exhibiting the best performance on the CUB dataset.
The results show that Softmax can well fit diverse classes while SVM is more suitable for classifying similar classes. 

Meta-ZSLs, including TAFE-Net, ZSML, and TGMZ, achieve the best performance among the algorithms in the same categories, indicating the effectiveness of incorporating meta-learning. Although most generative methods obtain worse performance than attribute-based methods on aPY, TGMZ's outperformance demonstrates the advantages of TAE in generative methods.

\subsection{Fusion Zero-shot Learning}


In this experiment, we use the fused dataset to validate the model's ability to handle datasets with diverse task distributions in the ZSL setting.
We omit to fuse AWA1 and AWA2, given their similarity.
Also, we omit to fuse the CUB dataset, considering CUB's attribute dim is much larger than the other datasets'.
Eventually, We claim three combined datasets: AWA1\&aPY, AWA2\&APY, and AWA1\&AWA2\&aPY (AWA\&aPY). To fit the ZSL setting, we apply zero-padding to transform attribute vectors to be of the same size. We combine the seen and unseen classes of source datasets as the new seen and unseen classes, separately, in the fused datasets.

Figure \ref{Fusion} shows the average per-class Top-1 accuracy results in the fusion-ZSL setting. 
Our model achieves the best performance among the compared algorithms, achieving 5.5\%, 7.9\%, and 3.6\% improvement on AWA1\&aPY, AWA2\&aPY, and AWA\&aPY, respectively.
The improvement demonstrates our model's capability to handle diverse task distributions and the advantages of task-aligned zero-shot learning.

\subsection{Generalized Zero-shot Learning}

\begin{figure}
\centering 
  \includegraphics[width=0.95\linewidth]{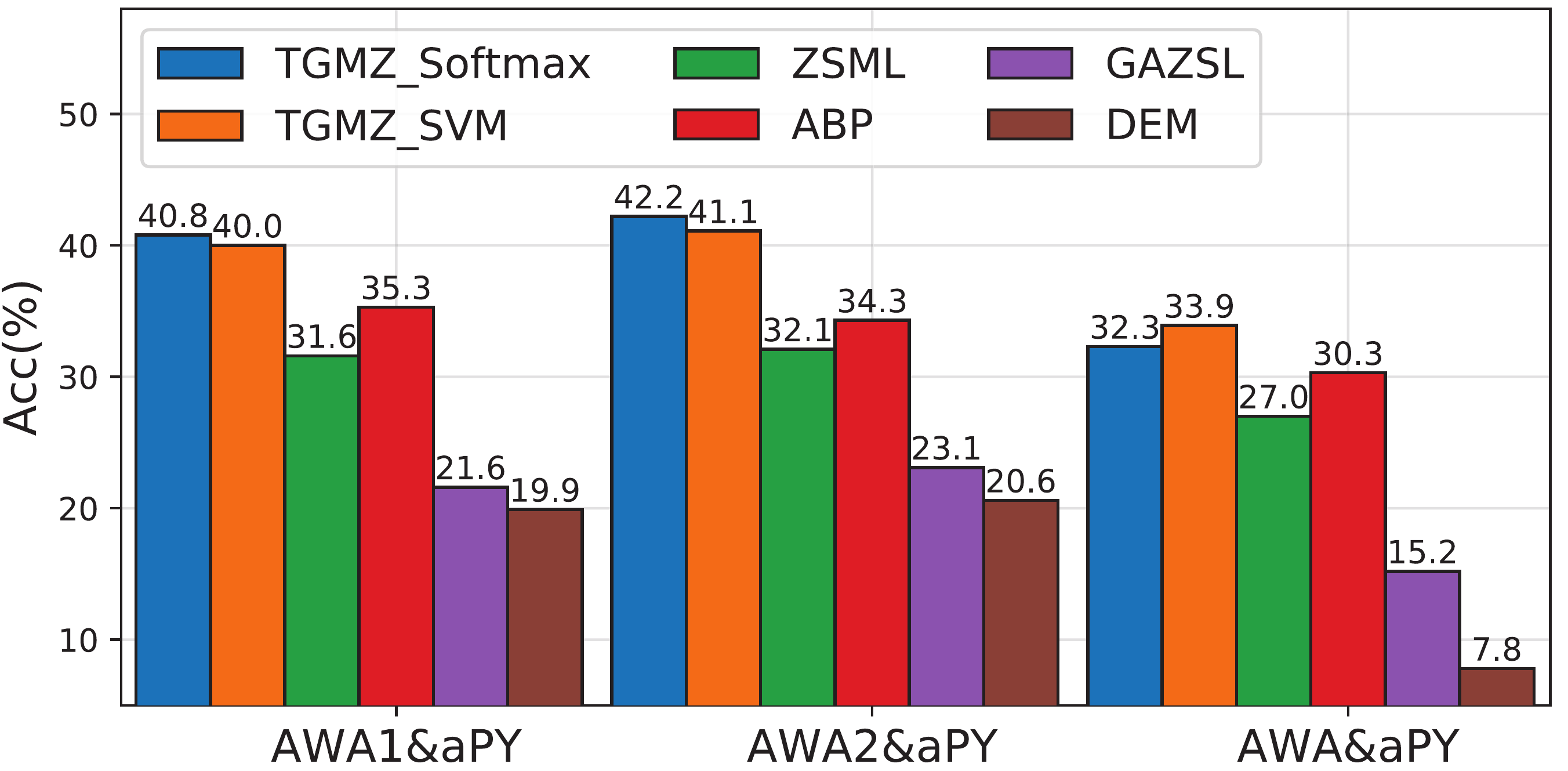}
    \centering
\caption{Fusion-ZSL average per-class Top-1 accuracies.}
\label{Fusion}
\end{figure}

We report our model's performance on four datasets using three evaluation metrics~\cite{xian2019zero}, namely average per-class Top-1 accuracy for unseen (U), seen (S), and the harmonic mean (H), where $H=\frac{2*U*S}{U+S}$. Compared with state-of-the-art results, our model yields 3.7\%, 2.8\%, and 2.5\% improvement in harmonic mean score on AWA2, CUB, and aPY, respectively. Also, our model obtains the best performance on the unseen classes of AWA1, AWA2, and CUB.
With respect to GZSL results, our model can effectively infer the potential visual-attribute correlation for unseen classes and prevent being biased towards the seen correlation.
Overall, our model exhibits consistent performance improvement in different settings (as shown in Table \ref{ZSL_experiment}, Table \ref{GZSL_experiment}, and Figure \ref{Fusion}), demonstrating the robustness and the superiority of TGMZ.

\begin{table*}
\centering
\caption{GZSL average per-class Top-1 accuracy results. * denotes meta-ZSL method. U and S represent the accuracy score for seen and unseen classes, respectively. H denotes the harmonic mean of U and S. $^{\psi}$ denotes using CNN-RNN feature for CUB dataset.}
\label{GZSL_experiment}
\small
\begin{tabular}{l|ccc|ccc|ccc|cccc}
\hline
\multirow{2}{*}{Method} & \multicolumn{3}{c}{AWA2} & \multicolumn{3}{c}{AWA1} & \multicolumn{3}{c}{CUB}& 
\multicolumn{3}{c}{aPY} \\ 
\cline{2-13}
& U & S & H & U & S & H & U & S & H & U & S & H\\
\hline
\hline

$^{\psi}$ESZSL~\cite{romera2015embarrassingly} & 5.9 & 77.8 & 11.0 & 6.6 & 75.6 & 12.1 & 12.6& 63.8 &21.0 &2.4 &70.1 &4.6 \\
$^{\psi}$LATEM~\cite{xian2016latent} & 11.5 & 77.3  & 20.0 & 7.3 & 71.7 & 13.3 & 15.2 & 57.3 & 24.0 & 0.1 & 73.0 & 0.2\\
$^{\psi}$SYNC~\cite{changpinyo2016synthesized}& 10.0 & 90.5 & 18.0 &8.9 &\textbf{ 87.3} & 16.2 & 11.5 & \textbf{70.9} & 19.8 & 7.4 & 66.3 & 13.3\\
$^{\psi}$DEM~\cite{zhang2017learning} & 30.5 & 86.4 & 45.1 & 32.8 & 84.7 & 47.3 & 19.6 & 57.9 & 29.2 & 11.1 & 75.1 & 19.4 \\
SAE~\cite{kodirov2017semantic} & 1.1 & 82.2 & 2.2 & 1.8 & 77.1 & 3.5 & 7.8 & 54.0 & 13.6 & 0.4 & \textbf{80.9} & 0.9\\
Gaussian-Kernal~\cite{zhang2018zero} & 18.9 & 82.7 & 30.8 & 17.9 & 82.2 & 29.4 & 21.6 & 52.8 & 30.6 & 10.5 & 76.2 & 18.5 \\
*TAFE-Net~\cite{Wang_2019_CVPR} &  36.7 & \textbf{90.6} & 52.2 & 50.5 & 84.4 & 63.2 & 41.0 & 61.4 & 49.2 & 24.3 & 75.4 & 36.8\\
APNet~\cite{liu2020attribute}  &  54.8 & 83.9 & 66.4 & 59.7 & 76.6 & 67.1 & 48.1 & 59.7 & 51.7 & 32.7 & 74.7 & 45.5\\
\hline
\hline
$^{\psi}$f-CLSWGAN~\cite{xian2018feature}  & 57.9 & 61.4 & 59.6 & 61.4 & 57.9 & 59.6 & 43.7 & 57.7 & 49.7 & -& -& -\\
GAZSL~\cite{zhu2018generative} & 35.4 & 86.9 & 50.3 &29.6 & 84.2 & 43.8 &31.7 & 61.3 & 41.8 &14.2 & 78.6 & 24.0\\
$^{\psi}$SE-GZSL~\cite{kumar2018generalized}  & 58.3 & 68.1 & 62.8 & 56.3 & 67.8 & 61.5 & 41.5 & 53.3 & 46.7 & - &- &- \\
GDAN~\cite{huang2019generative} & 32.1 & 67.5 & 43.5 & - & - & - & 39.3 & 66.7 & 49.5 & 30.4 & 75.0 & 43.4\\
ABP~\cite{zhu2019learning} & 55.3 &72.6 &62.6 & 57.3 &67.1  &61.8 &47.0 &54.8 &50.6& - &- &-\\

ZVAE~\cite{gao2020zero} & 57.1 & 70.9 & 62.5 & 58.2& 66.8 & 62.3 & 43.6 & 47.9 & 45.5 & 32.0 & 52.2 & 39.7\\
$^{\psi}$*ZSML~\cite{verma2020meta}  & 58.9 & 74.6 & 65.8 & 57.4 & 71.1 & 63.5 & 60.0 & 52.1 & 55.7 & \textbf{36.3} & 46.6 & 40.9\\

\hline
\hline
$^{\psi}$*TGMZ  & \textbf{64.1} & 77.3 & \textbf{70.1} & \textbf{65.1}& 69.4&\textbf{67.2}& \textbf{60.3}& 56.8& \textbf{58.5}& 34.8 & 77.1 &\textbf{48.0}
\\
\hline
\end{tabular}
\end{table*}
\begin{figure*}[h]
    \centering 
\begin{subfigure}{0.23\textwidth}
  \includegraphics[width=\textwidth]{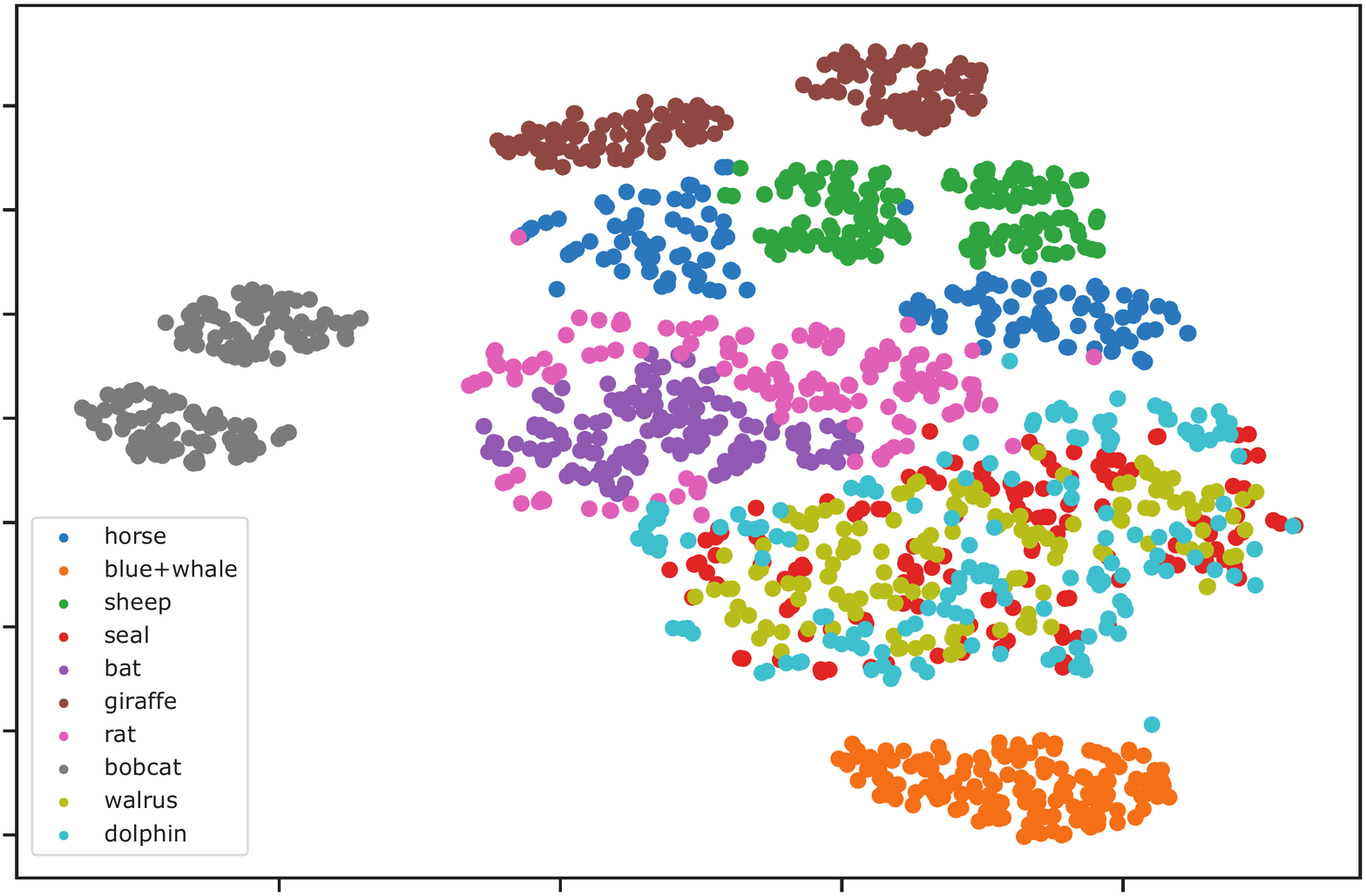}
    \centering
  \caption{ZSML's AWA2.}
\end{subfigure}\hfil 
\begin{subfigure}{0.23\textwidth}
  \includegraphics[width=\textwidth]{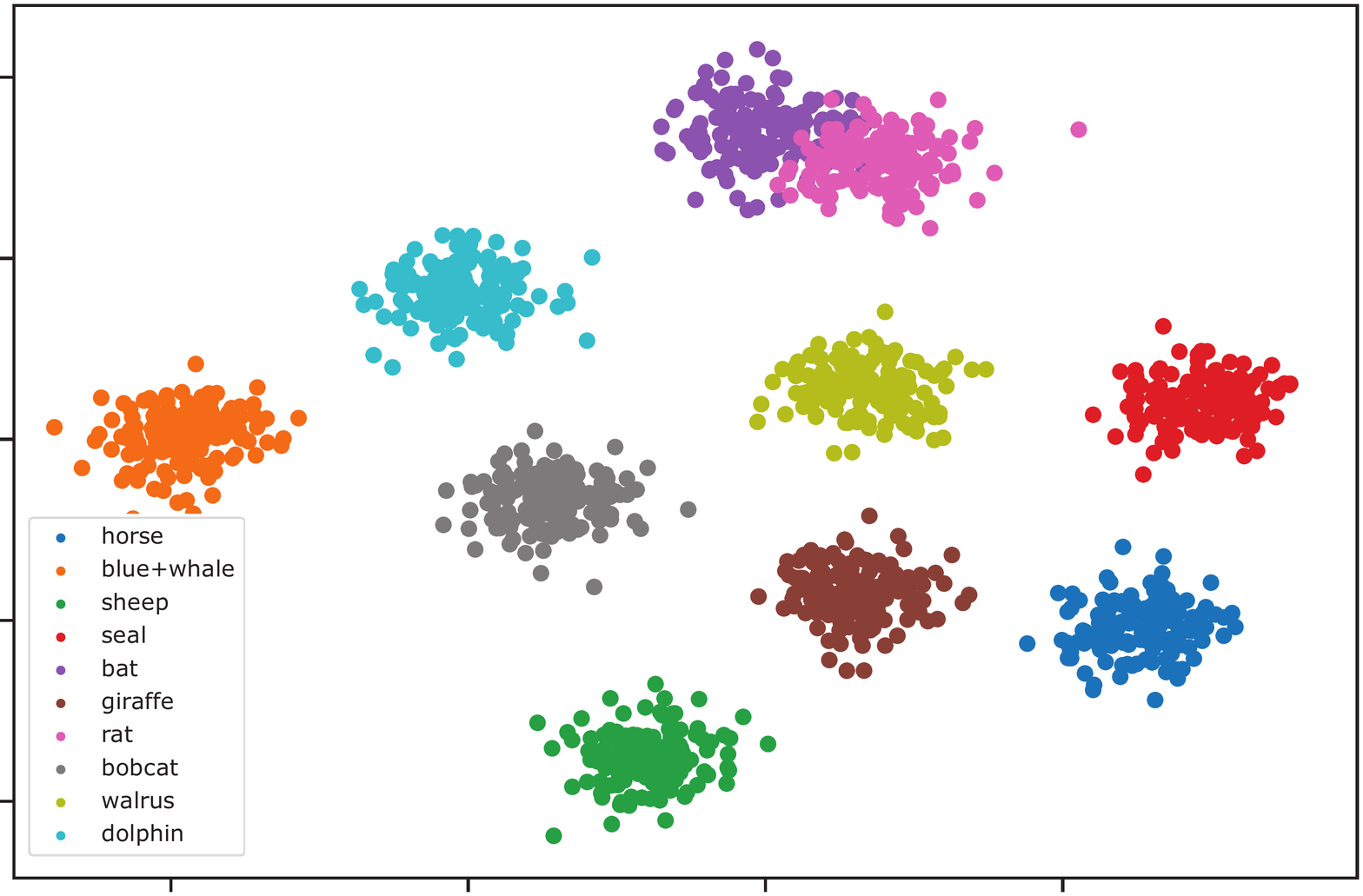}
    \centering
  \caption{Our AWA2.}
\end{subfigure}\hfil 
\begin{subfigure}{0.23\textwidth}
  \includegraphics[width=\textwidth]{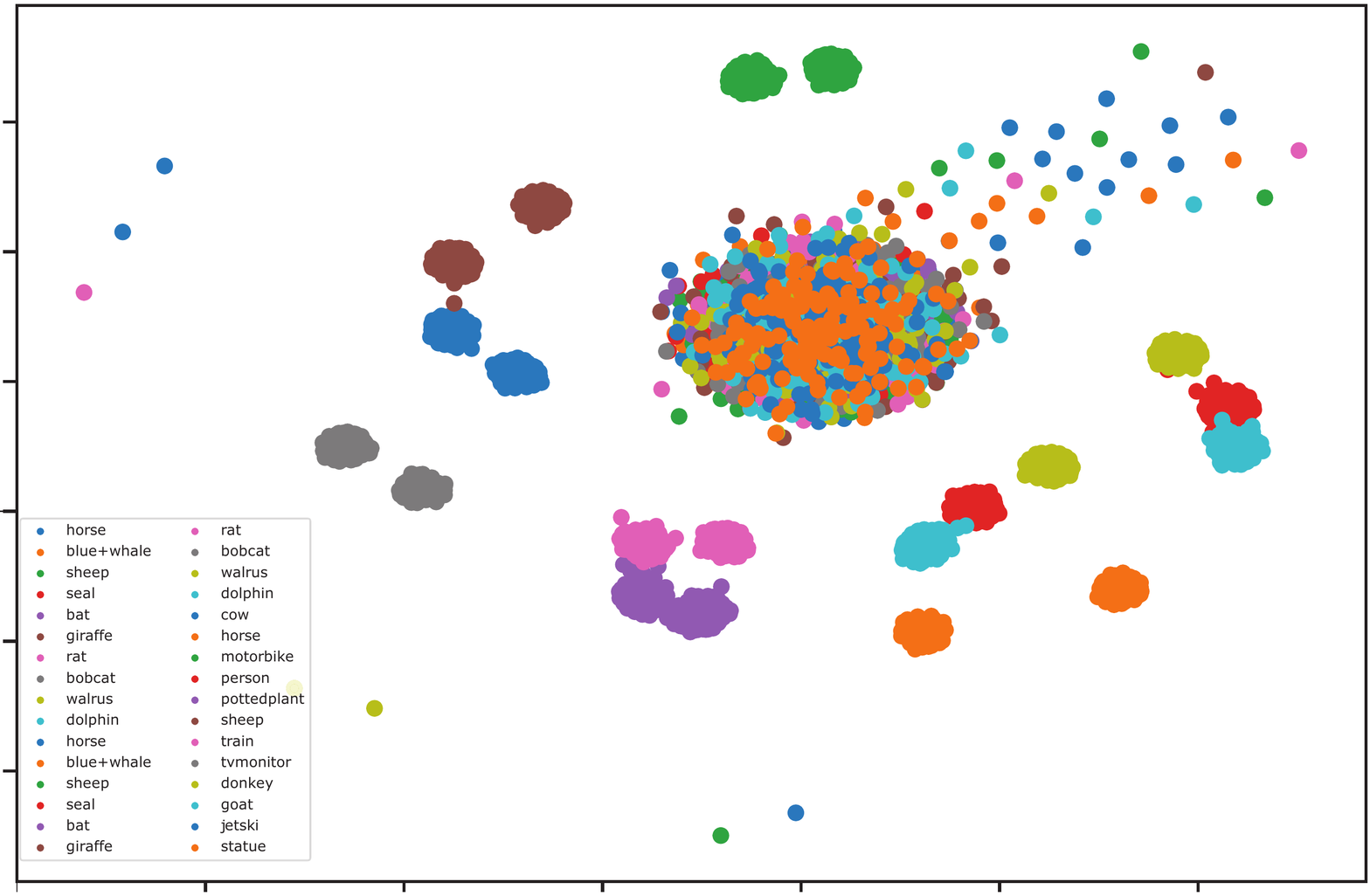}
    \centering
  \caption{ZSML's AWA\&aPY.}
\end{subfigure}\hfil 
\begin{subfigure}{0.23\textwidth}
  \includegraphics[width=\textwidth]{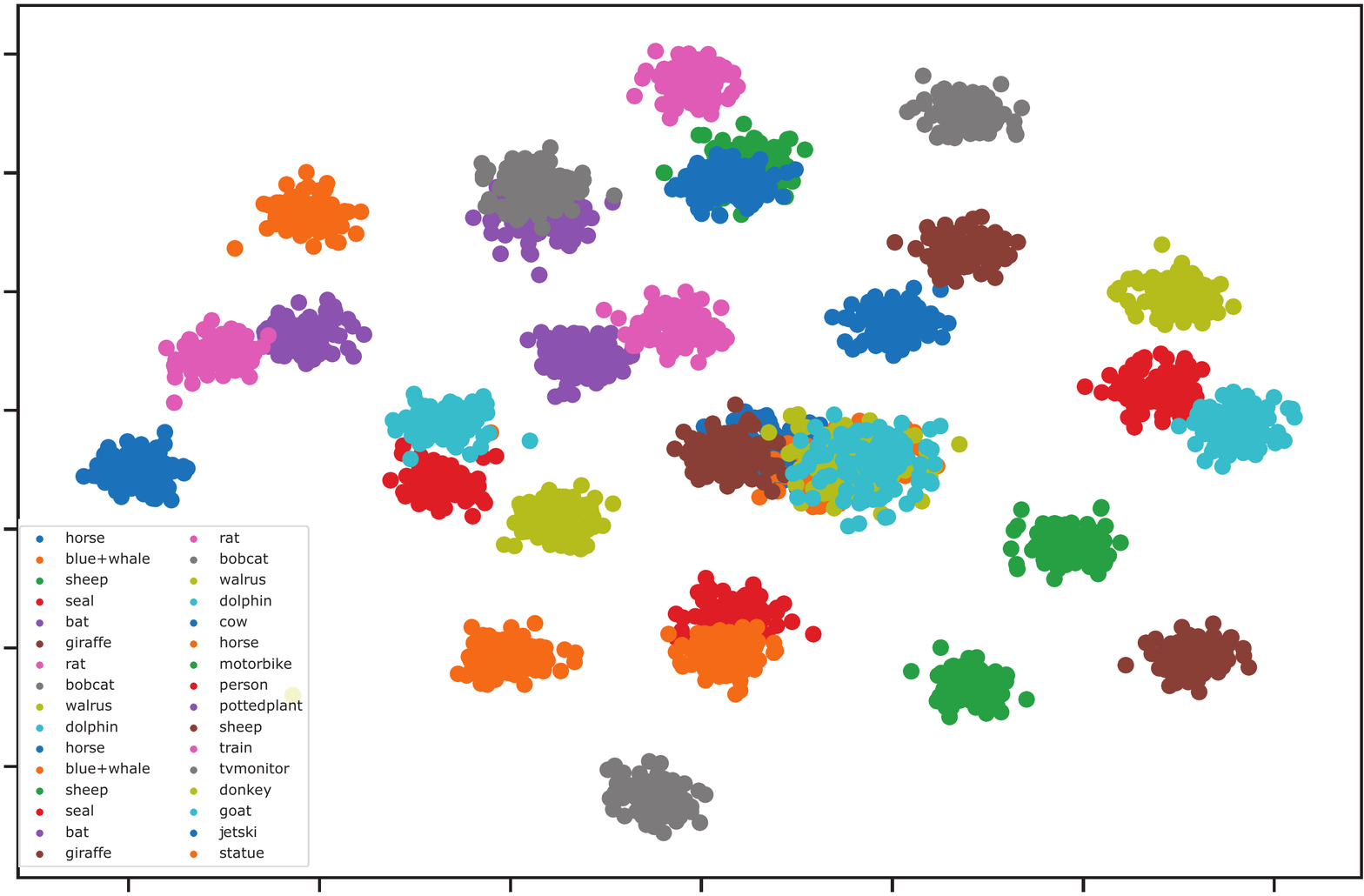}
    \centering
  \caption{Our AWA\&aPY.}
\end{subfigure}\hfil 
\caption{Synthetic feature embedding analysis.}
\label{embedding}
\end{figure*}

\begin{figure}
    \centering 
\begin{subfigure}{0.21\textwidth}
  \includegraphics[width=\textwidth]{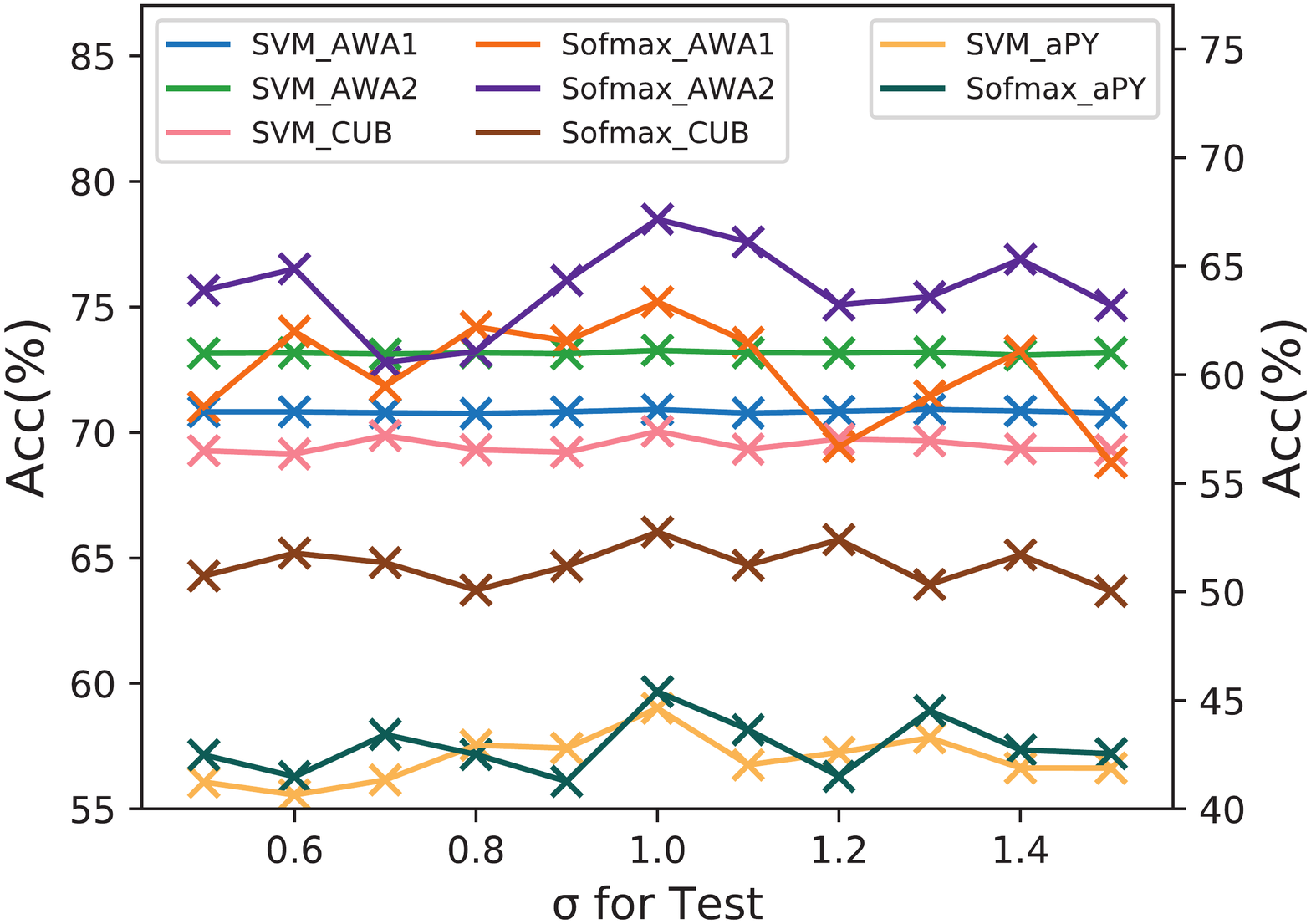}
    \centering
  \caption{Impact of $\sigma$. }
\end{subfigure}\hfil 
\begin{subfigure}{0.21\textwidth}
  \includegraphics[width=\textwidth]{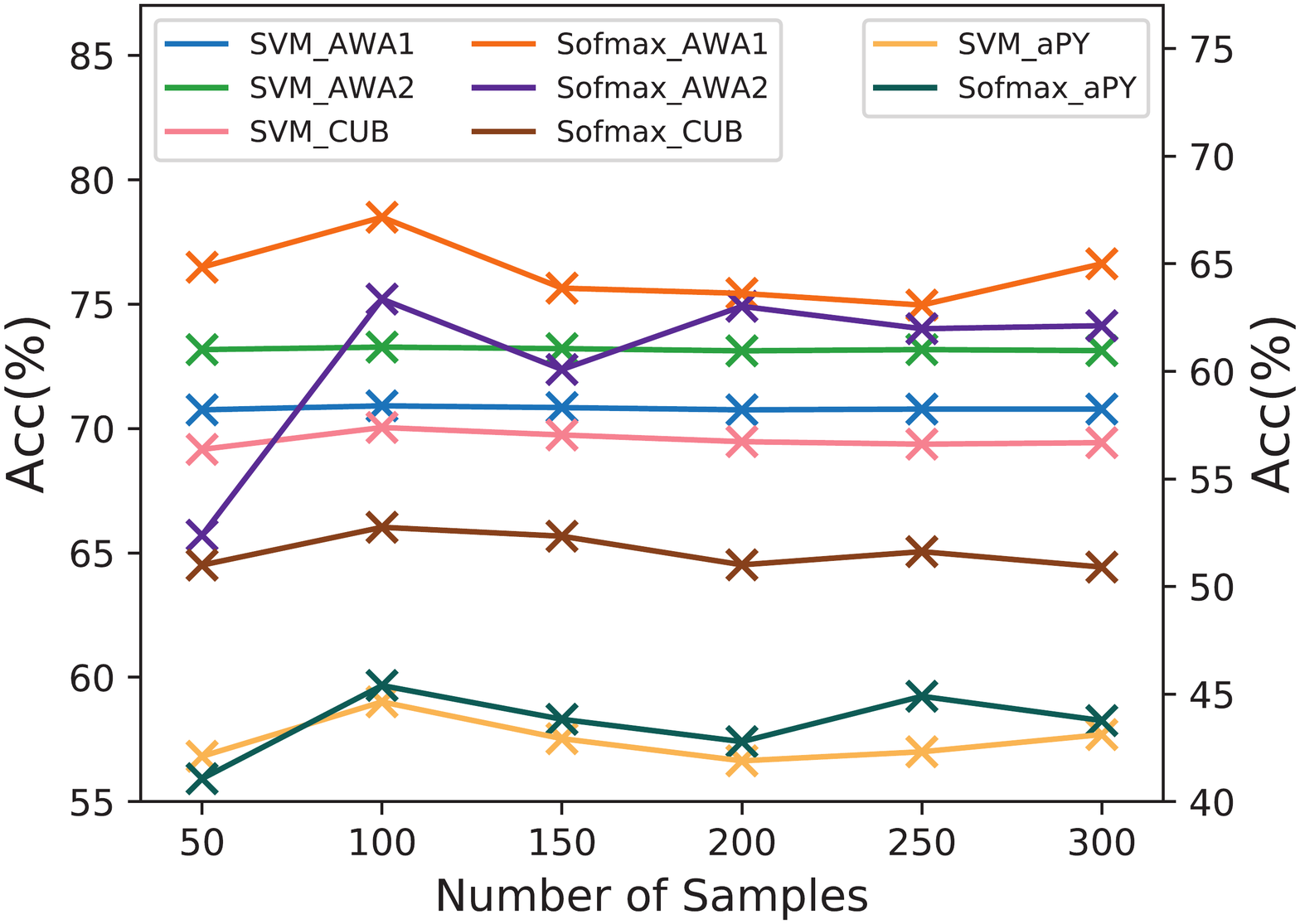}
    \centering
  \caption{Impact of Instance \#.}
\end{subfigure}\hfil 
\caption{Hyper-parameter analysis.}
\label{Hyperparameter}
\end{figure}

\subsection{Synthetic Feature Embedding Analysis}

Figure~\ref{embedding} provides a visualization of the synthetic features of ZSML and our model on AWA2 and the combined AWA\&aPY dataset. The original features are projected with t-Distributed Stochastic Neighbor Embedding (t-SNE).
ZSML can only generate discriminative distributions on a few classes in AWA\&aPY,  indicating it is biased towards some classes.
In contrast, our model can synthesize more discriminative feature space than ZSML on both AWA2 and AWA\&aPY. The discriminative embedding space demonstrates the effectiveness of our method in preventing meta-learner from being biased towards certain classes.

\subsection{Hyper-parameter Ablation Study}

We carry out the ablation study for $\sigma$ and the number of synthetic instances at the testing time. By default, we set $\sigma=1$ and the sample number to 100.
The ZSL results of SVM and Softmax on four datasets (Figure~\ref{Hyperparameter})
shows the parameter selection merely affect SVM on the four datasets, while
Softmax is only slightly affected by most parameters and is largely impaired when the sample number is 50 on AWA2.
Softmax achieves the best performance when $\sigma=1$ and gradually becomes better when provided a larger sample number. Thus, our model is generally robust on diverse parameters.

\section{Conclusion}
In this paper, we introduce a task-aligned meta generative model, TGMZ, to mitigate the potential biases in visual-attribute correlation learning for zero-shot learning. We propose a task-wise distribution alignment method to enhance the current zero-shot learning and evaluate our model on four popular benchmark datasets, demonstrating TGMZ's strong capability of learning diverse task distributions in three ZSL settings. To better illustrate the learned features, we visualize the data space of the synthetic features, which is more discriminative than the state-of-the-art generative meta-ZSL method in most classes. Overall, our method can optimize models in an unbiased way and achieve promising performance in image classification. In the future, we plan to extend TGMZ in more zero-shot learning scenarios, e.g., natural language processing and human activity recognition.

{\small
\bibliography{bio}}

 \end{document}